\documentclass{article}

\usepackage[nonatbib,final]{nips_2017}

\usepackage[utf8]{inputenc} % allow utf-8 input
\usepackage[T1]{fontenc}    % use 8-bit T1 fonts
\usepackage{hyperref}       % hyperlinks
\usepackage{url}            % simple URL typesetting
\usepackage{booktabs}       % professional-quality tables
\usepackage{amsfonts}       % blackboard math symbols
\usepackage{nicefrac}       % compact symbols for 1/2, etc.
\usepackage{microtype}      % microtypography
\usepackage{graphicx} % to add figures
\usepackage{placeins}

\usepackage[margin=1cm]{caption}
\usepackage{color}
\usepackage{tikz}
\usetikzlibrary{arrows, automata, backgrounds, calendar, chains,
decorations, hobby,intersections, matrix, mindmap, patterns, petri,
positioning, shadows, shapes, shapes.geometric,trees,shadings,
decorations.markings,snakes}

\usepackage{amsopn}

\newcommand{\X}{\mathcal{X}}
\newcommand{\Y}{\mathcal{Y}}
\newcommand{\Z}{\mathcal{Z}}

\newcommand{\D}{\mathcal{D}}
\newcommand{\C}{\mathcal{C}}

\newcommand{\x}{\vec{x}}

\renewcommand{\vec}[1]{\mathbf{#1}}
\newcommand{\p}{\vec{p}}
\newtheorem{remark}{Remark}

\title{Forward Thinking: Building Deep Random Forests}

\author{
Kevin Miller, Chris Hettinger, Jeffrey Humpherys, Tyler Jarvis, and David Kartchner\\
Department of Mathematics\\
Brigham Young University\\
Provo, Utah 84602 \\
\texttt{millerk5@byu.edu, hettinger@math.byu.edu, jeffh@math.byu.edu,}\\
\texttt{jarvis@math.byu.edu, david.kartchner@math.byu.edu}
}

\begin{document}

\maketitle

\begin{abstract}
The success of deep neural networks has inspired many to wonder whether other learners could benefit from deep, layered architectures.   We present a general framework called \emph{forward thinking} for deep learning that generalizes the architectural flexibility and sophistication of deep neural networks while also allowing for $(i)$ different types of learning functions in the network, other than neurons, and $(ii)$ the ability to adaptively deepen the network as needed to improve results.   This is done by training one layer at a time, and once a layer is trained, the input data are mapped forward through the layer to create a new learning problem.  The process is then repeated,  transforming the data through multiple layers, one at a time, rendering a new dataset, which is expected to be better behaved, and on which a final output layer can achieve good performance.  In the case where the neurons of deep neural nets are replaced with decision trees, we call the result a \emph{Forward Thinking Deep Random Forest} (FTDRF).  We  demonstrate a proof of concept by applying FTDRF on the MNIST dataset.  We also provide a general mathematical formulation, called \emph{Forward Thinking} that allows for other types of deep learning problems to be considered.
\end{abstract}

\section{Introduction}

Classification and regression trees are a fast and popular class of methods for supervised learning. For example, random forests \cite{breiman_01}, extreme gradient boosted trees \cite{ChenG16}, and conditional trees \cite{cond_trees} have consistently performed well in several benchmarking studies where various methods compete against each other \cite{StallkampSSI12, Statnikov2008, comparo}.

In recent years, however, deep neural networks (DNNs) have become a dominant force in several areas of supervised learning, most notably in image, speech, and natural language recognition problems, where deep learning methods are also consistently beating humans \cite{Krizhevsky,socher:2013}.  Although the use of multiple layers of neurons in a ``deep'' architecture has been well-known for many years \cite{Lecun98}, it wasn't until the discovery of feasible means of training via backpropagation that neural networks became successful.  However, DNNs still suffer from a variety of problems.   In particular, it is extremely expensive computationally to use backpropagation to train multiple layers of nonlinear activation functions \cite{LeCun:1998}.  This not only requires lengthy training, but also uses large quantities of memory, making the training of medium-to-large networks infeasible on a single CPU.  Moreover, DNNs are highly prone to overfitting and thus require both large amounts of training data and careful use of regularization to generalize effectively.  Indeed, the computational resources required to fully train a DNN are in many cases orders of magnitude more than other machine learning methods such as decision tree methods which perform almost as well on many tasks and even better on other tasks \cite{comparo}.  In other words, a lot of work is required to get at best only slightly better performance using DNNs.

In spite of these drawbacks, DNNs have out-performed simpler structures in a number of machine learning tasks, with authors citing the use of ``deep'' architectures as a necessary element of their success \cite{resnets}.  By stacking dozens of layers of weak learners (neurons), DNNs can capture the intricate relationships necessary to effectively solve a wide variety of problems.  Accordingly, we propose a generalization of the DNN architecture where neurons are replaced by other classifiers.  In this paper, we consider networks where each layer is a type of random forest, with neurons composed of the individual decision trees and show how such networks can be quickly trained layer-by-layer instead of paying the high computational cost of training a DNN all at once.

Random forests \cite{breiman_01,Murphy:2012} use ensembling of bootstrapped decision trees as weak classifiers, reporting the average or maximum value across all of the trees' outputs for classification probabilities and regression values.   In \cite{Liu_2008}, Liu notes that variety in individual weak learners is essential to the success of ensemble learning.  Accordingly, we use a combination of random decision trees and extra random trees \cite{Geurts2006} in each layer to increase variety and thus improve performance. We create both the random decision trees and extra random trees using the implementations provided in \texttt{Scikit-Learn} \cite{scikit-learn}.

It is important to note that Zhou and Feng \cite{ZhouF17} very recently posted a related idea called gcForest, where the layers of the deep architecture are comprised of multiple random forests.  In their network, the connections to subsequent layers are the outputs of random forests, whereas in our paper the outputs of the individual decision trees are passed to subsequent layers of decision trees.  In other words, they pass the results of the random forest through to the next layer ($4$ full random forests, each consisting of $1,000$ decision trees), whereas we pass the results of $2,000$ individual decision trees forward.  We get comparable results with less trees, but a higher memory requirement given that we are mapping more data to the next layer.  In particular with the MNIST dataset, they have an accuracy of 98.96\% and we get an essentially equivalent accuracy of 98.98\%.

In both our work and Zhou and Feng's work, these decision tree networks can be trained efficiently without the use of backpropagation. Each layer remains static once trained, and so the training data can be pushed through to train the next layer. Hence, the training time for a multi-layer forest should be much faster than training time for a traditional DNN architecture. We note that in a companion paper
\cite{ft_nets}
% (citation redacted for review)
, that we can also train a DNN in a similar fashion, without the use of backpropagation, thus also speeding up the training process.

The results of both gcForest and our study are convincing, and we believe that both papers confirm the validity of exploring deep architectures with decision trees and random forests.

In Section \ref{sec:math-description}, we give a careful description of the general architecture for a forward thinking deep network.  In Section \ref{architecture} we describe how the general theory is applied in the specific case of a Forwarding Thinking Deep Random Forest (FTDRF).  Details relating to the processing of data and the experimental results are in the subsequent sections.

\section{Mathematical description of forward thinking} \label{sec:math-description}

The main idea of forward thinking is that neurons can be generalized to any type of learner and then, once trained, the input data are mapped forward through the layer to create a new learning problem.  The process is then repeated,  transforming the data through multiple layers, one at a time, rendering a new data set, which is expected to be better behaved, and on which a final output layer can achieve good performance.

\subsection*{The input layer}

The data $\D^{(0)} = \{(\x_i^{(0)},y_i)\}^N_{i=1} \subset \X^{(0)}\times \Y$ are given as the set of input values $\vec{x}_i^{(0)}$ from a set $\X^{(0)}$ and their corresponding outputs $y_i$ in a set $\Y$.

In many learning problems, $\X^{(0)} \subset \mathbb{R}^d$, which means that there are $d$ real-valued features. If the inputs are images, we can stack them as large vectors where each pixel is a component.  In some deep learning problems, each input is a stack of images. For example, color images can be represented as three separate monochromatic images, or three separate channels of the image.

For binary classification problems, the output space can be taken to be $\Y= \{-1,1\}$.  For multi-class problems we often set $\Y= \{1,2,\ldots,K\}$.

\subsection*{The first hidden layer}

Let $\C^{(1)} = \{ C_1^{(1)}, C_2^{(1)},\ldots, C_{m_1}^{(1)}\}$ be a set of $m_1$ learning functions, $C_i^{(1)}:\X^{(0)}\to \Z_i^{(1)}$, for some codomain $\Z_i^{(1)}$ with parameters $\theta_i^{(1)}$.  This layer of learning functions (or learners) can be regression, classification, or kernel functions and can be thought of as defining new features. Let $\X^{(1)} = \Z_1^{(1)}\times \Z_2^{(1)}\times\cdots\times \Z_{m_1}^{(1)}$ and transform the inputs $\{\vec{x}_i^{(0)}\}^N_{i=1}\subset\X^{(0)}$ to $\X^{(1)}$ according to the map
\[
\vec{x}_i^{(1)} = (C_1^{(1)}(\vec{x}^{(0)}_i), C_2^{(1)}(\vec{x}^{(0)}_i),\ldots, C_m^{(1)}(\vec{x}^{(0)}_i)) \subset \X^{(1)},\quad i=1,\ldots,N.
\]
This gives a new data set $\mathcal{D}^{(1)} = \{(\vec{x}_i^{(1)},y_i)\}^N_{i=1}\subset \X^{(1)} \times \Y$.

In many learning problems $\Z^{(1)} = [-1,1]$, in which case the new domain $\X^{(1)}$ is a hypercube $[-1,1]^{m_1}$.  It is also common for $\Z^{(1)} = [0,\infty)$, in which case $\X^{(1)}$ is the $m_1$-dimensional orthant $[0,\infty)^{m_1}$.

The goal is to choose $\C^{(1)}$ to make the new dataset ``more separable,'' or better-behaved, than the previous dataset.  As we repeat this process iteratively, the data should become increasingly better-behaved so that in the final layer, a single learner can finish the job.

\subsection*{Additional hidden layers}

Let $\C^{(\ell)} = \{ C_1^{(\ell)}, C_2^{(\ell)},\ldots, C_{m_\ell}^{(\ell)}\}$ be a set (layer) of $m_\ell$ learning functions $C_i^{(\ell)}:\X^{(\ell-1)}\to \Z^{(\ell)}$.  This layer is again trained on the data $\D^{(\ell-1)} = \{(\x_i^{(\ell-1)},y_i)\}$. This would usually be done in the same manner as the previous layer, but it need not be the same; for example, if the new layer consists of different kinds of learners, then the training method for the new layer might also need to differ.

As with the first layer, the inputs $\{\x_i^{(\ell-1)}\}^N_{i=1}\subset \X^{(\ell-1)} = \Z_1^{(\ell-1)}\times \Z_2^{(\ell-1)}\times\cdots\times \Z_{m_{\ell-1}}^{(\ell-1)}$ are transformed to a new domain $\{\x_i^{(\ell)}\}^N_{i=1} \subset \X^{(\ell)} = \Z_1^{(\ell)}\times \Z_2^{(\ell)}\times\cdots\times \Z_{m_\ell}^{(\ell)}$ according to the map
\[
\x_i^{(\ell)} = (C_1^{(\ell)}(\x_i^{(\ell-1)}), C_2^{(\ell)}(\x_i^{(\ell-1)}),\ldots, C_{m_\ell}^{(\ell)}(\x_i^{(\ell-1)})),\quad i=1,\ldots,N.
\]
This gives a new dataset $\mathcal{D}^{(\ell)} = \{(\x_i^{(\ell)},y_i)\}^N_{i=1}\subset \X^{(\ell)} \times \Y$, and the process is repeated.

\subsubsection*{Final layer}

After passing the data through the last hidden layer,  we train the final layer, which consists of a single learning function $C_F:\X^{(n)}\to \Y$  on the data set  $\mathcal{D}^{(n)} = \{(\x_i^{(n)},y_i)\}^N_{i=1}\subset \X^{(n)}\times\Y$ to determine the outputs, where $C_F(\x_i^{(n)})$ is expected to be close to $y_i$ for each $i$.

\begin{remark}
While in this paper we have applied the multi-layer architecture of neural networks to decision trees in random forests, we note that this can be generalized to other types of classifiers. Where the decision trees in our architecture are analogous to the neurons in a DNN, other classifiers such as SVMs, gradient boosted trees, etc., should be able to be substituted for neurons in a similar fashion.
%(Isn't this discussed later?) The sequential training of layers of ensembled classifiers whose outputs are class probabilities of some form that are sent to the subsequent layer as input to train another ensemble of classifiers is our generalized concept. Just as Zhou and Feng suggest that their Cascading Forest architecture is building ``towards an alternative'' to DNN structures, their contributions along with our work here support the further exploration of this idea.
\end{remark}

\section{Forward thinking deep random forest architecture} \label{architecture}

In this section we describe the method of construction for layers of the Forwarding Thinking Deep Random Forest (FTDRF) architecture. We note the similarities to the routine termed CascadeForest in \cite{ZhouF17} and address these in this section.

\subsection{Multilayer random forests}\label{sec:ftdrf-description}

Using the notation of the previous section, we have training data $\D^{(0)} = \{\x_i^{(0)}, y_i\}_{i=1}^N$ (inputs and labels), where $\mathbf{x}_i^{(0)}$ are feature vectors and $y_i \in \{1,2, \ldots , K\}$ are the corresponding labels. An FTDRF consists of multiple layers $\C^{(1)}, \dots, \C^{(n)}$ of classifiers, where each layer $\C^{(\ell)}$ consists of a forest, comprised of a blend of random and extra random trees.

The output of each individual tree is a vector of class probabilities, as determined by the distribution of classes present in the leaf node into which the sample is sorted.  Specifically, given any decision tree, each leaf of the tree is assigned a vector of class probabilities, $\p=(p_1,\dots, p_K)$, corresponding to the proportion of training data assigned by the tree to the leaf in each class.

Each layer $\C^{(\ell)}$ is trained on the data $\D^{(\ell-1)} = \{\x_i^{(\ell-1)},y_i\}$, and $\x_i^{(\ell)}$ is the result of pushing the input $\x_i^{(\ell-1)}$ through that layer.
Specifically, for each input $\x_i^{(\ell-1)}$, the output of tree $j$ in layer $\ell$ is a probability vector $\p_j^{(\ell)}(\x_i^{(\ell-1)}) =  (p_{j,1}^{(\ell)}(\x_i^{(\ell-1)}), \dots, p_{j,K}^{(\ell)}(\x_i^{(\ell-1)}))$.  And these are concatenated together at each layer, so that for each
input $\x_i^{(\ell-1)}$, the output of layer $\C^{(\ell)}$ is an $m_\ell$-tuple of probability vectors, where $m_\ell$ is the number of trees in layer $\ell$.

 All such outputs for all trees in the layer are concatenated together to be the output of the layer for the given sample. This is done for all of the training data, hence transforming the data to be of dimension $K \times m_{\ell}$, where $K$ is the number of classes for the training dataset and $m_{\ell}$ is the number of trees in the current layer.

The outputs of each layer become the inputs to the next, until the data have been mapped through the final layer $\C^{(n)}$.  The final class prediction is made by averaging all the class probability output vectors from the $m_n$ decision trees in $\C^{(n)}$, and predicting the class with the highest probability. One could, of course, use any classifier to find an optimal combination of the weights for the final layer, but we do not explore this possibility in this paper.

\begin{figure}[htb]
  \includegraphics[scale=0.4]{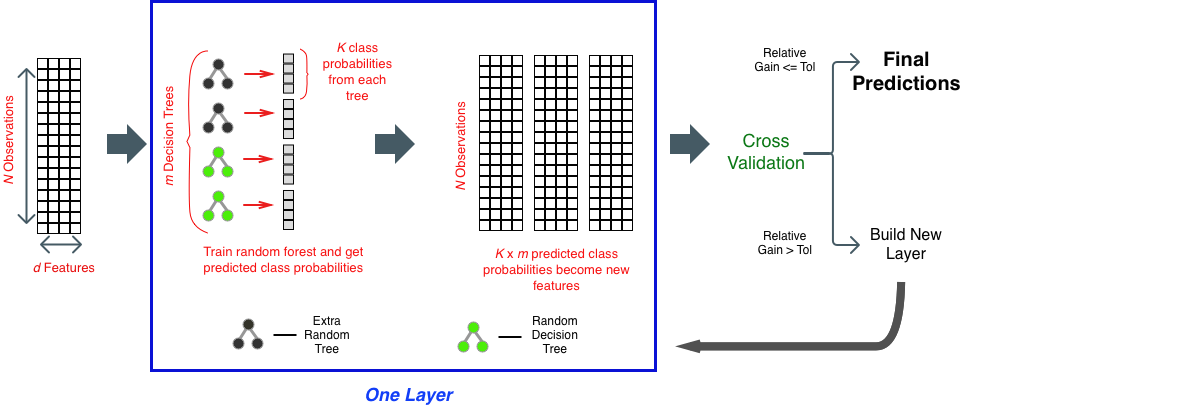}
  \caption{Forward Thinking Deep Random Forest (FTDRF)}
  \label{fig:ftdrf}
\end{figure}

\subsection{gcForest comparison}
Unlike Zhou and Feng's architecture for gcForest, our deep architecture of decision trees only requires the previous layer's output. In \cite{ZhouF17}, each layer passes both the class probabilities predicted by the random forests (not the individual decision trees) and the original data to each subsequent layer. Our model, on the other hand, passes only the output of the previous layer of individual decision trees to the next layer, to reduce the spatial complexity of network training and testing. Moreover, FTDRF seems to need fewer trees in each layer.  For example, in our FTDRF described in Section \ref{results} we obtained results comparable to \cite{ZhouF17} on MNIST, but we use only $2,000$ decision trees in each layer, whereas \cite{ZhouF17} uses 4 random forests of $1,000$ trees each (or $4,000$ trees per layer). Another distinction is that our final routine uses information gain entropy to calculate node splits, whereas gcForest implements gini impurity. We also ran some tests with gini impurity to determine node splits, but found that entropy usually performed better.

\subsection{Half-half random forest layers}
As is standard in random forests, a node split in a given decision tree is determined from a random subset containing $\sqrt{d}$ features of the input data passed to the layer. In a given layer, the collection of decision trees representing the layer contains both random decision trees, as well as extra random trees to introduce more variety into the layer. This is similar to the layers of \cite{ZhouF17}, where of the $4$ random forests in a given layer, $2$ of them are completely random forests \cite{Liu_2008}, closely related to extra random forests. An extra random forest increases tree randomization by choosing a random splitting value for each of the $\sqrt{d}$ features subset to determine the node split. In our scheme, we randomly assign trees to be of this type based on a Bernoulli draw of $p = 0.5$.

\subsection{Adding layers}

An advantage of forward thinking is that the total number of layers is determined by the data, rather than by a human designer.  In the case of FTDRF, the choice of whether to create a new layer or terminate is determined by a cross validation scheme.  After each layer is constructed, we evaluate the accuracy on a holdout sample consisting of $20\%$ of the training data to determine the relative gain produced by the last added layer.  If the layer meaningfully increased the validation accuracy (i.e., the relative gain is above a chosen threshold), then we proceed and add another layer to the FTDRF.  Once the relative gain of a new layer falls below the threshold, we stop adding new layers and obtain predictions via the final layer of our network.  For our network, we chose a relative gain threshold of $1\%$.  Some results are shown in Table~\ref{table:no_mgs}, below.

We note that the trees in the layers of our specific implementation here were  not created using boosting (e.g. XGBoost \cite{ChenG16}), but we expect that doing so could be beneficial and possibly lead to increased accuracy.

%%TJ this was a complete duplication of earlier discussion, hence removed.
%\subsection{Mapping data through layer}
%%TODO: get notation to match other section
%Once the addition of another layer is decided upon from the cross-validation training, the input data is mapped through the decision tree classifiers, as described in Section~\ref{sec:ftdrf-description} in the layer and the class probabilities in the leaf nodes are the outputs. Instead of the layer deciding the average class or regression value for a given sample, the output of each individual tree is a vector of class probabilities, as determined by the distribution of classes present in the leaf node that the sample is sorted into. All such outputs for all trees in the layer are concatenated together to be the output of the layer for the given sample. This is done for all of the training data, hence transforming the data to be of dimension $K \times n_{trees}$, where $K$ is the number of classes for the training dataset and $n_{trees}$ is the number of trees in the current layer.
%

\section{Preprocessing of image data}

The decision tree structure of FTDRF requires sufficient training data to avoid overfitting in the first few layers.
% It seems that the layering of the random forest decision trees quickly overfits to the training data if the number of training samples and dimension of the data aren't large enough.
The state-of-the-art algorithms for dealing with image data in classification use preprocessing and transforming techniques, such as convolutions.  Accordingly, we experimented with two of these techniques for FTDRF: a single-pixel ``wiggle'' and multi-grained scanning (MGS).

\subsection{Single-pixel wiggle}

\begin{figure}[htb]
\centering
  \includegraphics[scale=0.4]{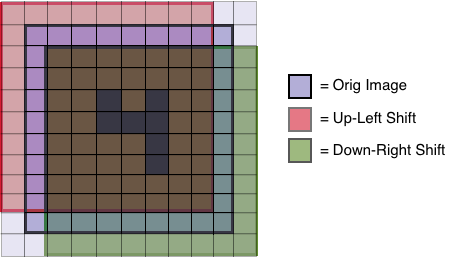}
  \caption{Single pixel wiggle visualization}
  \label{fig:sp}
\end{figure}

For the data set used here, we augmented the training data by a single pixel ``wiggle'' technique. That is, for each training image in the MNIST training set, we include copies of the images shifted around in $4$ diagonal directions (up-left, up-right, down-left, and down-right) by one pixel, see Figure \ref{fig:sp}. This data augmentation yields the results seen  in Table \ref{table:no_mgs}. A further way to augment the feature representation of the images is presented in the following Section \ref{mgs}, via a routine called Multi-Grained Scanning \cite{ZhouF17}.

\subsection{Multi-grained scanning (MGS)}\label{mgs}

 In \cite{ZhouF17}, a scheme similar to convolution is proposed, termed Multi-Grained Scanning (MGS), which we implemented for the FTDRF architecture.
We use the exact same process that Zhou and Feng do in their MGS scheme \cite{ZhouF17} so as to be able to compare the results of our architecture in the subsequent network structure FTDRF. We view this MGS process as a preprocessing transformation akin to the convolutions of convolutional neural networks (CNNs), with the benefits and strengths that such transformations provide.

\begin{figure}[htb]
  \includegraphics[scale=0.4]{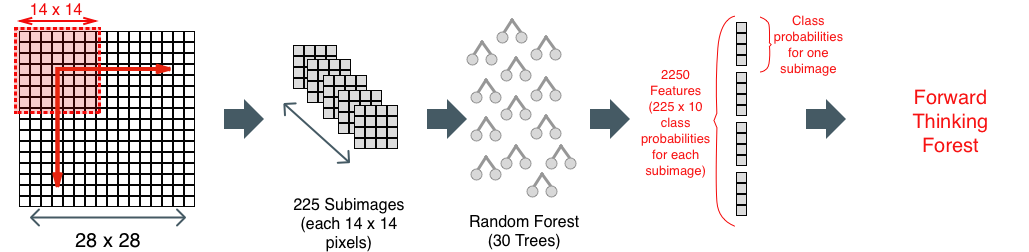}
  \caption{Multi-grained scanning (MGS) routine, window size = 14}
  \label{fig:mgs}
\end{figure}

In MGS, windows of a set number of sizes are obtained inside the training set images (for the MNIST dataset window sizes are $7 \times 7$, $9 \times 9$, and $14 \times 14$).

For a given window size, the corresponding windows contained inside of all training images are used as a training set to construct a random forest and an extra random forest whose outputs are the class probabilities. Unlike our routine for the building of the FTDRF layers that output the class probabilities determined by each individual decision tree in the layer, this scheme outputs the class probabilities determined by the whole random forest. Hence, for a given window size, the output of the random forest for each image window is a vector of the class probabilities. For all samples fed through these random forests, the outputs of all image windows are concatenated together to produce a feature vector representing classification probabilities of each of the windows (see Figure \ref{fig:mgs}).

With the $3$ window sizes specified, the outputs of each of the $2$ random forests for the respective window sizes are all concatenated together. This feature vector is the new representation of each given sample fed through the MGS process. With this transformation of the training data (and subsequently the testing data), we train the FTDRF layers as previously described in Section \ref{architecture}.

%\section{FTDRF implementation}
%
%We have implemented the FTDRF and MGS routines in Python, using the \texttt{Scikit-Learn} \cite{scikit-learn} implementation for CARTs and random forests. Our code, in using these other packages isn't as optimized as a professional level package. Hence, our code is not as fast as it could be if focus were made on the speed of the implementation. As this is not the main focus of this paper in justifying this method, we note that the speed of decision tree construction in random forests will allow for fast algorithms to be produced in this deep network structure.
%
%Results of this implementation are described below.

\section{FTDRF results on MNIST}\label{results}

We present results for an FTDRF on the MNIST handwriting digit recognition dataset, where each sample is a $(28\times28)$ black and white image of isolated digits, written by different people. The dataset is split into a training set with $60,000$ (see note below) samples and testing set of $10,000$ samples.

\subsection{Results with single-pixel wiggle}

For each training image, we created $4$ more $(28\times28)$ images via the single-pixel wiggling technique to augment the size of the training data. The layers of FTDRF contained $2,000$ decision trees ($\sim1,000$ random decision trees and $\sim1,000$ extra random trees). Layers were grown until the the relative gain was less than $1\%$, totaling $3$ layers. Node splits were determined by calculating the information gain entropy. We cite our results and the results of Zhou and Feng \cite{ZhouF17} to compare, as their architecture is most relevant to ours.  We note, however, that Zhou and Feng do not augment data in this test. The results are:
\begin{table}[h!]

\centering
 \begin{tabular}{||c | c | c||}
 \hline
  \textbf{Model}& \textbf{\# Trees} & \textbf{Accuracy} \\
 \hline
 gcForest & 4000 & 97.85\% \\
 \hline
 FTDRF & 2000 & 97.58\% \\
 \hline
 \end{tabular}
 \vskip0.2cm
 \caption{MNIST results without MGS}
 \label{table:no_mgs}
\end{table}

\subsection{Results with MGS}

Table~\ref{table:MGS} presents the results of our architecture compared to Zhou's gcForest \cite{ZhouF17}, including the MGS preprocessing routine. Note then that we do \emph{not} augment the dataset with the single pixel augmentation as we did previously. In this test, window sizes of 7, 9, and 14 were used for the MGS step, creating a total of $6$ random forests ($3$ random forests and $3$ extra random forests) to transform the data for the FTDRF training. Then, training data was passed through to the FTDRF step, where layers consisted of $2,000$ decision trees ($\sim1,000$ random decision trees and $\sim1,000$ extra random trees) but in this step, only $2$ layers were necessary to achieve the desired relative validation error threshold. The results are:

\begin{table}[h]
\centering
 \begin{tabular}{||c | c | c||}
 \hline
\textbf{Model}& \textbf{\# Trees} & \textbf{Accuracy} \\
 \hline
 gcForest & 4000 & 98.96\% \\
 \hline
 FTDRF & 2000 & 98.98\% \\
 \hline
 FTDRF & 500 & 98.89\% \\
 \hline
 \end{tabular}
 \vskip0.2cm
\caption{MNIST results with MGS}
\label{table:MGS}
\end{table}

\FloatBarrier

\section{Related work}
As we have mentioned, the work of Zhou and Feng \cite{ZhouF17} is similar to our work, and their preprocessing technique of MGS was adapted for our use in testing. Our FTDRF primarily differs from Zhou and Feng's gcForest in that gcForest passes the outputs of whole random forests concatenated onto the original data to each subsequent layer, whereas we pass only the outputs of the individual trees to subsequent layers. The gcForest algorithm was very successful in a variety of classification settings, including image and sequential data (in which MGS is applied), along with other non-sequential data (in which MGS is not applied).

Another related idea is that of ``stacking'' classifiers \cite{Wolpert92, Zhou:2012}. In the context of \cite{Wolpert92}, an ensemble of classifiers is trained and then further improvements are made by adding a classifier or ensemble of classifiers to interpret the best way to combine the outputs of the original ensemble's classifiers. In the perspective of our deep architecture and its building process, the idea of stacking therefore could be compared to a $2$-layer architecture, with a relatively small second layer. Our idea proposes to continue the process, with larger layers stacked similarly to a DNN.

The connections between random forest and DNN structure were explored in \cite{biau:neuralRF, deep-neural-decision-forests, wolf:deep_neural_pre}. These papers assert that random forest construction bears similarity to DNN construction and that random forests can thus be transformed into neural networks and vice versa.  More specifically, the mathematical dependencies between DNN nodes have been shown to be similar to the dependencies between decision tree leaf nodes in random forests.
While these methods draw connections between the construction of decision trees in random forests and DNNs, our work is fundamentally different in the idea of ensembling decision trees together in layers resembling a DNN architecture. As explained in Section \ref{architecture}, our architecture represents mapping data through different hypercubes in hopes of iterating towards more easily classified data.

\subsection*{Reproducibility}
 All python code used to produce our results is available in our github repository at %\texttt{URL blinded for refereeing process}.
https://github.com/tkchris93/ForwardThinking.
\subsubsection*{Acknowledgments}

This work was supported in part by the National Science Foundation, Grant Number 1323785 and the Defense Threat Reduction Agency, Grant Number HDRTA1-15-0049.

%\newpage
\bibliographystyle{plain}
\bibliography{references}
\end{document}